\documentclass{article}

\PassOptionsToPackage{numbers, compress}{natbib}

    \usepackage[preprint]{neurips_2021}

\usepackage[utf8]{inputenc} 
\usepackage[T1]{fontenc}    
\usepackage{hyperref}       
\usepackage{url}            
\usepackage{booktabs}       
\usepackage{amsfonts}       
\usepackage{nicefrac}       
\usepackage{microtype}      
\usepackage{xcolor}         

\usepackage{subcaption} 
\usepackage{adjustbox}
\usepackage{multicol}
\usepackage{amsmath}

\title{Simon Says: Evaluating and Mitigating Bias in Pruned Neural Networks with Knowledge Distillation}

\author{%
Cody ~Blakeney \\
  Department of Computer Science\\
  Texas State University\\
  San Marcos, TX 78758 \\
  \texttt{cjb92@txstate.edu} \\
   \And
   Nathaniel Huish \\
   Department of Computer Science\\
   Texas State University \\
   San Marcos, TX 78758 \\
   \texttt{njh71@txstate.edu} \\
   \AND
   Yan Yan \\
   Department of Computer Science \\
   Illinois Institute of Technology\\ 
   Chicago, IL 60616 \\
   \texttt{yyan34@iit.edu} \\
   \And
   Ziliang Zong \\
   Department of Computer Science \\
   San Marcos, TX 78758 \\
   \texttt{ziliang@txstate.edu} \\
}

\begin{document}

\maketitle

\begin{abstract}

In recent years the ubiquitous deployment of AI has posed great concerns in regards to algorithmic bias, discrimination, and fairness. Compared to traditional forms of bias or discrimination caused by humans, algorithmic bias generated by AI is more abstract and unintuitive therefore more difficult to explain and mitigate. A clear gap exists in the current literature on evaluating and mitigating bias in pruned neural networks. In this work, we strive to tackle the challenging issues of evaluating, mitigating, and explaining induced bias in pruned neural networks. Our paper makes three contributions. First, we propose two simple yet effective metrics, Combined Error Variance (CEV) and Symmetric Distance Error (SDE), to quantitatively evaluate the induced bias prevention quality of pruned models. Second, we demonstrate that knowledge distillation can mitigate induced bias in pruned neural networks, even with unbalanced datasets. Third, we reveal that model similarity has strong correlations with pruning induced bias, which provides a powerful method to explain why bias occurs in pruned neural networks. Our code is available at \url{https://github.com/codestar12/pruning-distilation-bias} 

\end{abstract}

\section{Introduction}

Deep neural networks (DNNs) are widely deployed across many domains but they are incredibly compute- and memory-intensive. To solve this problem, compression techniques such as pruning and quantization have been widely used to reduce the model size, computation demand, and energy consumption of complex DNNs without compromising top-line metrics (e.g. top-1 and top-5 accuracy). Nevertheless, recent literature has exposed the prevalence of undesirable biases in compressed neural networks\cite{hooker2019compressed, hooker2020characterising}. For example, Hooker et al. found that pruned networks often sacrifice the accuracy of a subset of the classes to preserve overall accuracy in classification tasks \cite{hooker2019compressed}. This exacerbates existing issues of algorithmic bias already observed in DNNs like CV models working better for people with lighter skin \cite{buolamwini2018gender}.    Unfortunately, these algorithmic biases generated by AI are more abstract and unintuitive, which makes it harder to explain, evaluate, and mitigate. Despite the rising concerns in bias and fairness of AI, research on how to evaluate and mitigate bias is still in its infancy. 

In this paper, we strive to tackle the challenging issues of evaluating, mitigating, and explaining bias caused by pruning neural networks. Specifically, we make the following three contributions. 

\begin{itemize}
  \item \textbf{Bias evaluation}: We propose two simple yet effective metrics, Combined Error Variance (CEV) and Symmetric Distance Error (SDE), to quantitatively evaluate the bias prevention quality of pruned neural networks. Compared to existing metric such as Pruning Identified Exemplars (PIEs), CEV and SDE have clear advantages. First, they do not require training large populations of models. Two models can be compared directly. Second, SDE and CEV can evaluate any compression methods (like PIEs or CIEs) but they are more objective than PIEs or CIEs by providing clear measurement of model quality in bias prevention. 
  \item \textbf{Bias mitigation}: We demonstrate that knowledge distillation can effectively mitigate bias in pruned neural networks, even with unbalanced datasets. We apply state-of-the-art knowledge distillation algorithms on network pruning and use our proposed CEV and SDE metrics to evaluate their model bias. Our results clearly show that knowledge distillation provides a viable solution for mitigating bias in pruned neural networks. We also investigate the impact of unbalanced dataset on model bias. Although unbalanced datasets can amplify bias issues in pruned networks, our experiments confirm that knowledge distillation remain effective in mitigating bias in pruned models with unbalanced datasets. 
  
  
  \item \textbf{Bias explanation}: We reveal that model similarity has strong correlations with compression induced bias, which provides a powerful method to explain why bias occurs in pruned neural networks. We utilize Singular Vector Canonical Correlation Analysis (SVCCA) \cite{raghu2017svcca} to compare similarity of layer representations between the original (non-pruned) and pruned models. Our experimental results unanimously show that pruned models that yield high similarity to original models induce less bias.

\end{itemize}



\section{Evaluation Metrics for Model Bias}

To evaluate the severity of bias in different AI models, it is essential to define fair and effective metrics. Recently, Google has conducted important work by exposing the prevalence of undesirable biases in compressed neural networks \cite{hooker2020characterising} and proposed Pruning Identified Exemplars (PIEs) \cite{hooker2019compressed} as a metric for measuring compression induced bias in pruned neural networks. In this section, we discuss PIEs and its limitations. Meanwhile, we propose two new metrics, Combined Error Variance (CEV) and Symmetric Distance Error (SDE), and demonstrate their effectiveness by comparing to PIEs.

\subsection{Pruning Identified Exemplars}

\begin{figure*}[!h]
    \centering

    \begin{subfigure}[b]{0.23\paperwidth}
        \includegraphics[width=\textwidth]{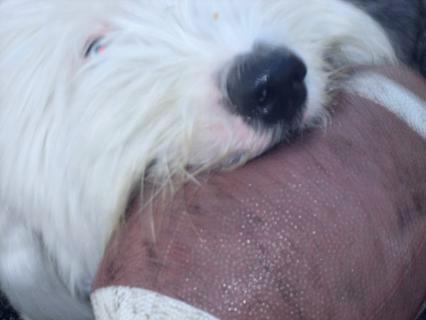}
        \caption{Shared PIE}
    \end{subfigure}
    \hfill
   \begin{subfigure}[b]{0.23\paperwidth}
        \includegraphics[width=\textwidth]{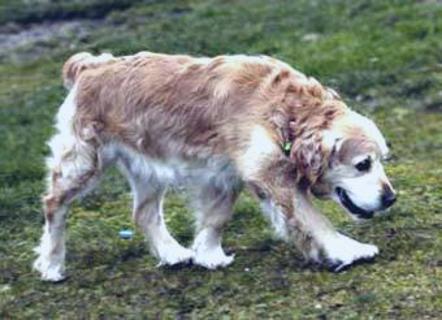}
        \caption{PIE without distillation}
    \end{subfigure} 

        \caption{PIEs may represent noisy or confusing data (e.g. image (a) is hard for human to classify).}
    \label{fig:PIE_ex}
\end{figure*}


PIEs are defined as the following equation, which represents images in a dataset for which pruning explicitly changes the behavior of the answer. In \cite{hooker2019compressed}, they trained a population of 30 models for each compression method and sparsity level. They then defined an image $i$ as an exemplar if modal label $Y_{i,t}^{M}$, or class most predicted by the \textit{t}-compressed model population disagrees with the label produced from the original unpruned networks. 

\[
PIE_{i,t} = \begin{cases}
1,  & \text{if } y_{i, 0}^M \neq y_{i, t}^M\\
0,  & \text{otherwise}
\end{cases}\
\]

While PIEs help reveal the bias issues in pruned models, not every image reported as a PIE represents a problem. A good portion of PIEs represent images that are equally hard for human to classify and may simply be a case where the uncompressed networks overfit to learn the example. Figure \ref{fig:PIE_ex}(a) is a PIE image that is found both in pruned models with and without distillation. This image doesn't have a clear subject and is hard for human to label as well. It is probably not proper to say that miss labeling this example is a poor model. However, Figure \ref{fig:PIE_ex}(b) is a very different case. The subject is in the center of the image, is not obstructed, and contains no other objects that might be labeled. Missing labeling it indicates low quality of pruned models. 

While PIEs are an important step towards identifying and detecting bias in pruned models, they have clear limitations. First, finding PIEs requires training a population of models for which to compare a modal answer. Therefore, it is not practical to directly compare the quality of a pruned model to the original model. Second, as PIEs only compare examples labeled by two populations (or original model), they cannot provide insights about distribution variance or shift, which are critical to evaluate model bias. 

To address the weaknesses of PIEs, we propose two new metrics, Combined Error Variance (CEV) and Symmetric Distance Error (SDE). Pruning and quantization have been observed to sacrifice accuracy on a subset of class in classification tasks in order to retain overall top-k accuracy \cite{hooker2019compressed}. To catch and reflect this bias, our metrics are designed to quantify both the \textit{spread} of the change in classification error as well as changes in \textit{how} the model is making mistakes. As a result, both of our proposed metrics consider the distribution of change in false positive and false negative rates (FPR, FNR) for all classes.

\subsection{Combined Error Variance}

The principle of the Combined Error Variance (CEV) metric is to discourage the bias of sacrificing one class for the benefit of another class. Therefore, CEV calculates the variance of the FNR/FPR changes and penalizes changes in FNR/FPR away from the model's average change. Mathematically, CEV is defined and calculated as follows.

Let $X_i$ be a tuple of the FPR and FNR for class $i$ of the compressed model and $\hat{X_i}$ be the original models FPR/FNR tuple. We first find a the normalized change in FPR/FNR $\delta{X_i}$ and the mean normalized change $\delta{X_{\mu}}$.

\begin{multicols}{2}
    \begin{equation} \label{norm_change}
        \delta{X_i} =  \frac{X_i - \hat{X_i}}{\hat{X_i}} \times 100
    \end{equation}
    \begin{equation}\label{mean_change}
        \delta{X}_{\mu} = \frac{1}{n}\sum_{i=0}^{n} (\delta{X_{i}})
    \end{equation}
\end{multicols}

The variance of the change in FPR and FNR is calculated for all $n$ classes as follows.

\begin{equation} \label{cev}
cev = \frac{1}{n}\sum_{i=0}^{n} (\delta{X}_{\mu} - \delta{X_{i}})^{2}
\end{equation}

\subsection{Symmetric Distance Error}

The principle of the Symmetric Distance Error (SDE) metric is to discourage another undesirable bias behavior in pruned models. That is, a class with more training examples or that has similar features to another class is more frequently to be chosen by the model as its capacity degrades. Conversely, classes with fewer training examples are more likely to be cannibalized by the model and guessed less frequently. To reflect this bias behavior, SDE calculates "how far away" from balanced is the change in FPR/FNR for a single class error. More intuitively, if we make a scatter plot with changes in FPR and FNR as X and Y values, the diagonal line in that plot would be a perfectly balanced change in FPR/FNR. Therefore, the SDE can be calculated as the symmetric distance of each change to that balance line.  


Mathematically, it can be proven that for a line given by the equations $ax + by + c = 0$, the distance $d$ from any point $(x_0, y_0)$ can be derived from the following equation: 
\begin{equation} \label{SDE_base}
    \centering
    d = \frac{|a(x_0) + b(y_0)|}{\sqrt{a^2 + b^2}}
\end{equation}

In our specific context, the diagonal of the Cartesian plane (i.e. the balance line) is $x = y$ or $x - y = 0$. Therefore, given any change in the form of FNR (i.e. x) and FPR (i.e. y) coordinate, the symmetric distance of that change to the balance line can be calculated as:  
    \begin{equation}
        \centering
        d = \frac{|(1)(x_0) + (-1)(y_0)|}{\sqrt{(1)^2 + (-1)^2}}  = \frac{|x_0 -y_0|}{\sqrt{2}}
    \end{equation}


Once the symmetric distance of each change is calculated, SDE of a pruned model can be calculated as the mean magnitude of normalized FP/FN rate difference divided by $\sqrt{2}$.  

\begin{equation}\label{sym_error}
sde = \frac{1}{n}\sum_{i=0}^{n}\frac{ | \delta{FNR_i} - \delta{FPR_i} |}{\sqrt{2} }
\end{equation}

\subsection{Comparison between CEV/SDE and PIEs}


To demonstrate the effectiveness of CEV and SDE for evaluating model bias in pruned models, we outline the following experiments and compare with the PIEs metric. 

\begin{table}[h]
\caption{Knowledge distillation methods and corresponding abbreviations}
\label{tab:kd}
\centering
\resizebox{.7\textwidth}{!}{
\begin{tabular}{lll}
\toprule
Distillation Method                          & Proposed By & Abbreviation \\
\midrule
Knowledge Distillation                       & Hinton et al. \cite{hinton2015distilling}    & KD          \\
Attention Transfer                           & Zagorukyo et al. \cite{Zagoruyko2017AT} & AT          \\
Similarity-Preserving Knowledge Distillation & Tung et al. \cite{tung2019similarity}      & SP          \\
Flow of Solution Procedure                   & Yim et al.  \cite{yim2017gift}     & FSP         \\
Probablistic Knowlege Transfer               & Passalis et al. \cite{pkt_eccv}  & PKT         \\
Contrastive Representation Distillation      & Tian et al.  \cite{tian2019contrastive}    & CRD         \\
\bottomrule
\end{tabular}}
\end{table}

\subsubsection*{Pruning Details}
We train a CNN ResNet \cite{he2016deep} (ResNet32x4) model on CIFAR100 \cite{Krizhevsky09learningmultiple} and prune it to various sparsities using Filter-wise Structured Pruning. We use pruning as the baseline model and the rest of models are pruned jointly with one of the state-of-the-art knowledge distillation methods shown in Table \ref{tab:kd}. We utilize Tian et al's \cite{tian2019contrastive} implementation for all distillation methods tested. For all experiments the models are pruned initially to 10\% sparsity then gradually pruned every 5 epochs until the desired sparsity is reached according to the AGP \cite{zhu2017prune} schedule. All pruning is completed at the halfway point of training and allowed to continue to finetune for another 120 epochs. The chosen sparsity levels are 30\%, 45\%, 60\%, 75\%. Each layer in the model is pruned to the same sparsity. We do not perform any layer sensitivity analysis or prune layers at different ratios. Although that may have resulted in higher accuracy, our goal is not to reach state-of-the-art compression ratios but study the effect of pruning and how to mitigate bias.

\begin{figure*}[!h]
    \centering
    \includegraphics[width=.97\textwidth]{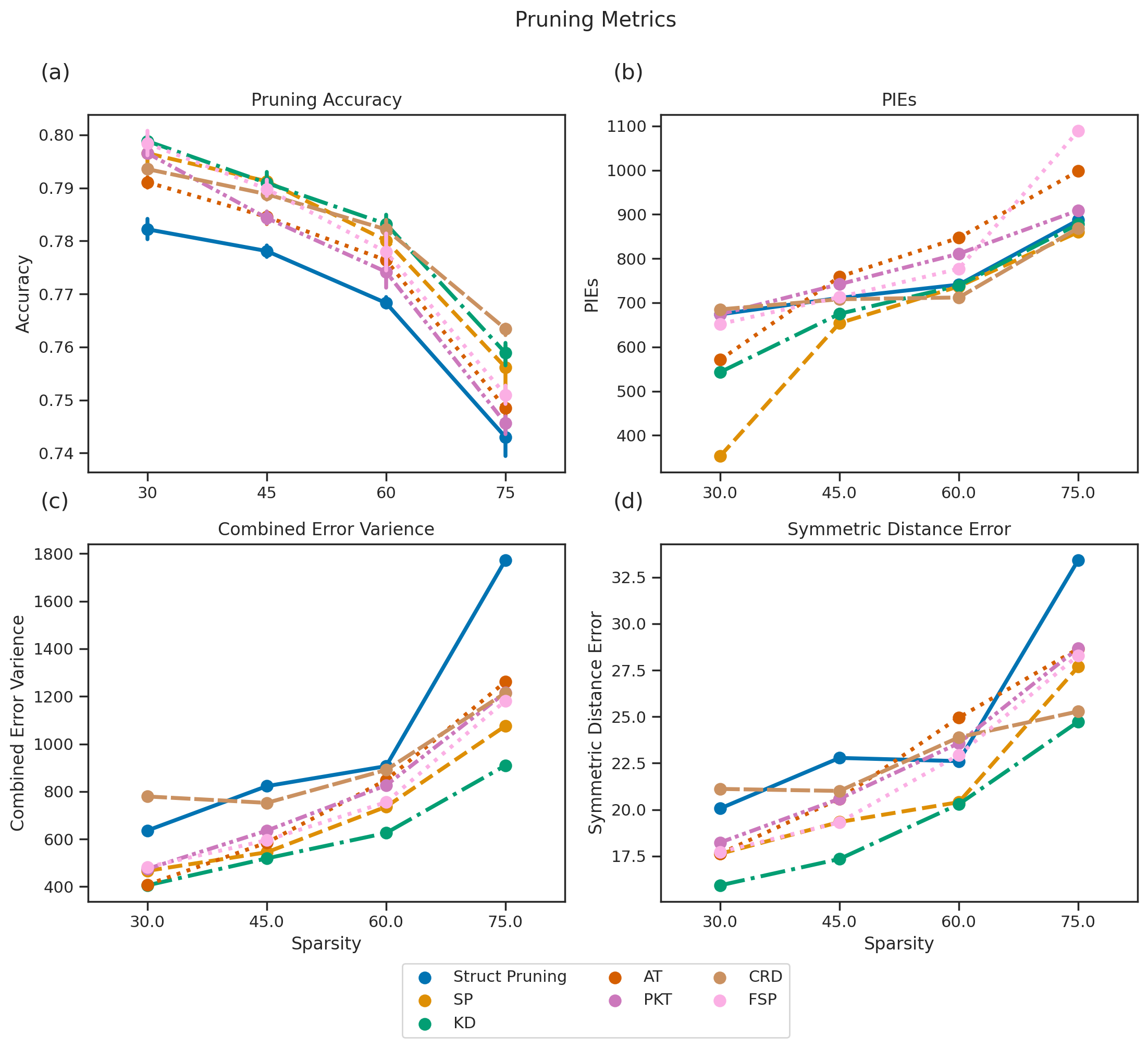}

    \caption{Accuracy, PIEs, CEV, and SDE results and comparison}
    \label{fig:curves}
\end{figure*}

\subsubsection*{Results}
 
\begin{figure}[h]
     \centering
     \begin{subfigure}[b]{0.44\textwidth}
         \centering
         \includegraphics[width=\textwidth]{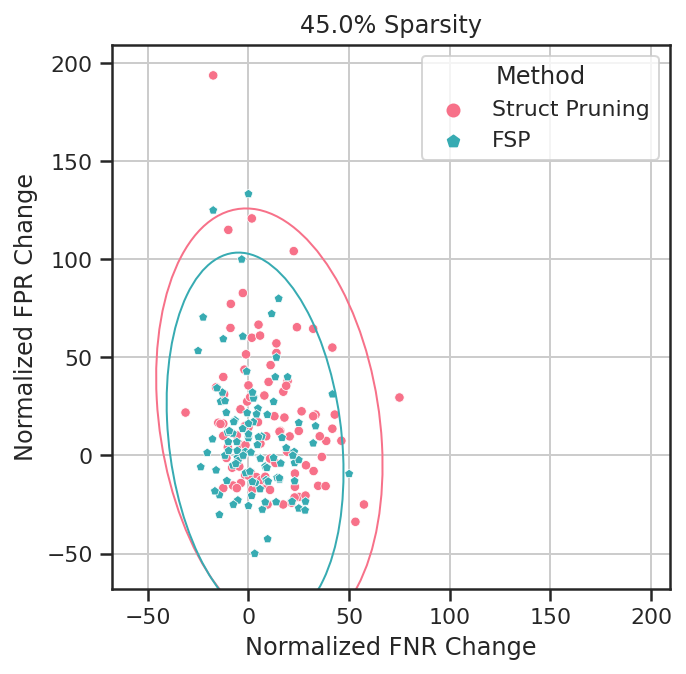}
        \caption{Pruning vs. FSP}
        \label{fig:PvFSP_cifar100}
     \end{subfigure}
     \hfill
     \begin{subfigure}[b]{0.44\textwidth}
         \centering
         \includegraphics[width=\textwidth]{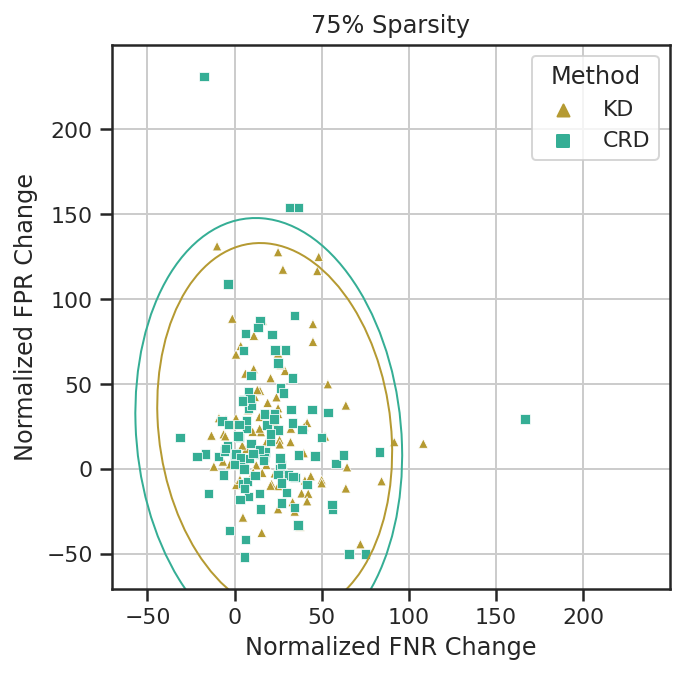}
        \caption{KD vs. CRD}
        \label{fig:KDvCRD_cifar100}
     \end{subfigure}

    \caption{Scatter plot of normalized false positive and false negatives rate (FPR/FNR) change for different distillation methods on CIFAR100. Each point represents a single class's normalized change. Ellipses contain 95\% of data points.}
    \label{fig:FNR/FPR_CIFAR}
    
\end{figure}

 Figure \ref{fig:curves} plots the experimental results on accuracy and model bias of each pruned model evaluated using the PIE metric and our CEV/SDE metrics. We can observe from Figure \ref{fig:curves} (a) that several methods reach nearly identical accuracy (e.g. KD, SP, and FSP at 45\% sparsity). Figure \ref{fig:curves} (b) shows the evaluation results of model bias using the PIEs metric, from which we can see that the ranking of three models with similary accuracy would be SP > KD > FSP (i.e. less PIEs indicates better models). However, the ranking will change dramatically to KD > FSP > SP (i.e. small CEV/SDE indicates better models) when evaluating using CEV (ref. Figure \ref{fig:curves} (c)) and SDE (ref. Figure \ref{fig:curves} (d)) respectively. To find out which metrics better reflect the bias issues in pruned models, we plot the actual distribution of FPR/FNR change of the baseline pruned model and FSP in \ref{fig:FNR/FPR_CIFAR} (a). We can immediately see that the ellipse of FSP is much smaller than the ellipse of SP. Since ellipses contain 95\% of data points, it clearly shows the error distribution of FSP is certainly better than SP (i.e. less biased), although FSP has a higher number of PIEs. 
 

 An even more exaggerated case is the comparison of the CRD and KD models. CRD has a higher accuracy by nearly half a percentage point at 75\% sparsity (ref. Figure \ref{fig:curves} (a)) and the number of PIEs for the two models at this sparsity is nearly identical (ref. Figure \ref{fig:curves} (b)). CRD seems to be a better model. However, when plotting the error distribution of KD and CRD at 75\% sparsity (ref. Figure \ref{fig:FNR/FPR_CIFAR} (b)), we can clearly observe that KD has better error distribution for the majority of classes and its outliers are much less pronounced than the CRD distilled models. These important characteristics are missed by PIEs but correctly reflected by our CEV/SDE metrics (ref. Figures \ref{fig:curves} (c) and (d)). These two examples provide strong evidence that our metrics are more descriptive than the PIEs metric in evaluating model bias. 





\section{Mitigating Bias with Knowledge Distillation}




From previous experiments, we notice that all pruned models with distillation achieves higher accuracy at every sparsity than the model pruned without distillation (ref. Figure \ref{fig:curves} (a)). Surprisingly, we also notice that knowledge distillation can effectively mitigate pruning induced bias by tightening the distribution between the normalized change in FPR and FNR (ref. Figure \ref{fig:FNR/FPR_CIFAR} (a) as an example). It appears that distilled pruned models have tighter distribution of classes overall and their extreme outliers are less pronounced.

\subsection*{Data Induced Bias}


\begin{table}[!ht]
\centering
\caption{CIFAR100 class sample rate and mean class top-1 accuracy for original dataset (Org) and down sampled dataset (Bias) pruned at 45\% sparsity.}
\resizebox{.95\columnwidth}{!}{%
\begin{tabular}{llrrrrrrrrrr}

\toprule
        & Method & \multicolumn{2}{c}{ Struct Pruning} & \multicolumn{2}{c}{KD} & \multicolumn{2}{c}{ FSP} & \multicolumn{2}{c}{ FSP + KD} & \multicolumn{2}{c}{ SP} \\
        & {} &            Org &  Bias &   Org &  Bias &   Org &  Bias &      Org &  Bias &   Org &  Bias \\
Class Name & Percentage Remain &                &       &       &       &       &       &          &       &       &       \\
\midrule
apple & 10.00 &          88.74 & \textcolor{red}{66.99} & 91.24 & 79.24 & 90.74 & 72.99 &    91.49 & 78.74 & 91.49 & 75.49 \\
bicycle & 10.00 &          84.24 & \textcolor{red}{33.24} & 88.74 & 62.24 & 85.74 & 37.74 &    85.74 & 58.24 & 85.99 & 42.74 \\
crocodile & 10.00 &          92.49 & \textcolor{red}{63.99} & 92.49 & 79.49 & 92.74 & 71.99 &    92.24 & 79.49 & 92.49 & 71.24 \\
maple tree & 10.00 &          88.74 & \textcolor{red}{55.99} & 87.24 & 77.99 & 89.24 & 63.49 &    89.99 & 77.24 & 91.24 & 63.74 \\
shrew & 10.00 &          76.49 & \textcolor{red}{29.99} & 80.24 & 45.49 & 78.74 & 34.99 &    74.99 & 48.99 & 75.74 & 34.99 \\
butterfly & 20.00 &          83.24 & \textcolor{red}{59.74} & 84.24 & 71.24 & 82.99 & 68.49 &    85.24 & 69.49 & 83.99 & 65.74 \\
can & 20.00 &          89.74 & 80.99 & 88.49 & 81.24 & 89.74 & 80.49 &    89.24 & 82.74 & 90.24 & 81.24 \\
oak tree & 20.00 &          77.74 & \textcolor{red}{63.74} & 78.49 & 70.74 & 77.49 & 65.49 &    79.49 & 68.74 & 77.24 & 68.49 \\
palm tree & 20.00 &          88.24 & \textcolor{red}{71.99} & 90.24 & 79.49 & 90.49 & 77.99 &    90.49 & 80.24 & 90.24 & 77.49 \\
train & 20.00 &          95.49 & 84.99 & 94.74 & 90.24 & 94.24 & 86.49 &    95.99 & 87.24 & 94.99 & 86.24 \\
dinosaur & 50.00 &          92.74 & 89.74 & 92.74 & 90.49 & 92.74 & 88.99 &    91.74 & 89.99 & 92.99 & 90.24 \\
elephant & 50.00 &          92.24 & \textcolor{red}{86.99} & 89.74 & 89.74 & 93.49 & 90.74 &    91.99 & 89.49 & 91.74 & 89.74 \\
flatfish & 50.00 &          82.74 & \textcolor{red}{74.49} & 86.49 & 80.99 & 86.49 & 76.24 &    86.49 & \textbf{82.24} & 85.99 & 76.49 \\
orange & 50.00 &          81.49 & \textcolor{red}{70.49} & 82.49 & 75.99 & 78.24 & 72.74 &    81.74 & 75.99 & 82.49 & 74.24 \\
raccoon & 50.00 &          75.49 & \textcolor{red}{62.74} & 77.99 & 72.49 & 77.99 & 68.24 &    78.99 & 72.24 & 79.49 & 70.74 \\
\bottomrule

\toprule
        & Method & \multicolumn{2}{c}{SP + KD} & \multicolumn{2}{c}{PKT} & \multicolumn{2}{c}{PKT + KD} & \multicolumn{2}{c}{AT} & \multicolumn{2}{c}{AT + KD} \\
        & {} &     Org &  Bias &   Org &  Bias &      Org &  Bias &   Org &  Bias &     Org &  Bias \\
Class Name & Percentage Remain &         &       &       &       &          &       &       &       &         &       \\
\midrule
apple & 10.00 &   92.24 & 79.24 & 90.74 & 79.99 &    90.49 & 83.99 & 89.74 & 81.24 &   89.74 & \textbf{84.99} \\
bicycle & 10.00 &   88.24 & 60.74 & 85.74 & 49.99 &    88.24 & \textbf{69.49} & 87.24 & 59.74 &   86.99 & 68.99 \\
crocodile & 10.00 &   92.99 & 81.24 & 90.99 & 71.99 &    92.49 & 84.74 & 90.99 & 73.99 &   92.99 & \textbf{86.24} \\
maple tree & 10.00 &   91.24 & 75.74 & 88.49 & 65.99 &    88.74 & \textbf{82.49} & 87.49 & 72.99 &   88.99 & 82.49 \\
shrew & 10.00 &   76.99 & 48.99 & 79.49 & 41.24 &    80.74 & 54.74 & 73.99 & 43.49 &   81.24 & \textbf{56.24} \\
butterfly & 20.00 &   84.24 & 73.49 & 86.99 & 72.99 &    87.24 & 75.74 & 88.74 & 72.49 &   87.49 & \textbf{77.24} \\
can & 20.00 &   88.74 & 83.49 & 87.24 & 81.74 &    86.99 & 85.99 & 89.49 & 83.24 &   89.74 & \textbf{87.24} \\
oak tree & 20.00 &   78.24 & 70.74 & 77.24 & 69.24 &    78.24 & 71.49 & 75.74 & 69.49 &   77.24 & \textbf{71.99} \\
palm tree & 20.00 &   90.24 & 80.99 & 89.99 & 77.49 &    89.99 & 83.99 & 90.74 & 81.74 &   88.24 & \textbf{84.99} \\
train & 20.00 &   95.49 & 90.24 & 93.99 & 84.99 &    95.49 & 90.49 & 96.24 & 86.49 &   93.74 & \textbf{90.74} \\
dinosaur & 50.00 &   92.99 & 91.49 & 93.99 & 88.49 &    92.74 & \textbf{92.74} & 92.74 & 90.74 &   93.74 & 91.74 \\
elephant & 50.00 &   89.99 & 89.74 & 91.99 & 90.49 &    92.99 & \textbf{91.99} & 92.24 & 89.49 &   90.99 & 90.99 \\
flatfish & 50.00 &   85.99 & 79.24 & 84.49 & 76.99 &    84.24 & 80.49 & 81.49 & 76.99 &   82.49 & 76.74 \\
orange & 50.00 &   81.74 & 74.49 & 85.49 & 76.24 &    84.24 & \textbf{77.99} & 84.24 & 76.24 &   84.24 & 76.74 \\
raccoon & 50.00 &   77.74 & 74.24 & 79.24 & 72.49 &    78.24 & \textbf{74.74} & 75.24 & 68.24 &   78.99 & 73.74 \\
\bottomrule
\end{tabular}
}
\label{tab:sampling}

\end{table}

\begin{figure}[h]
     \centering
     \begin{subfigure}[b]{0.44\textwidth}
         \centering
         \includegraphics[width=\textwidth]{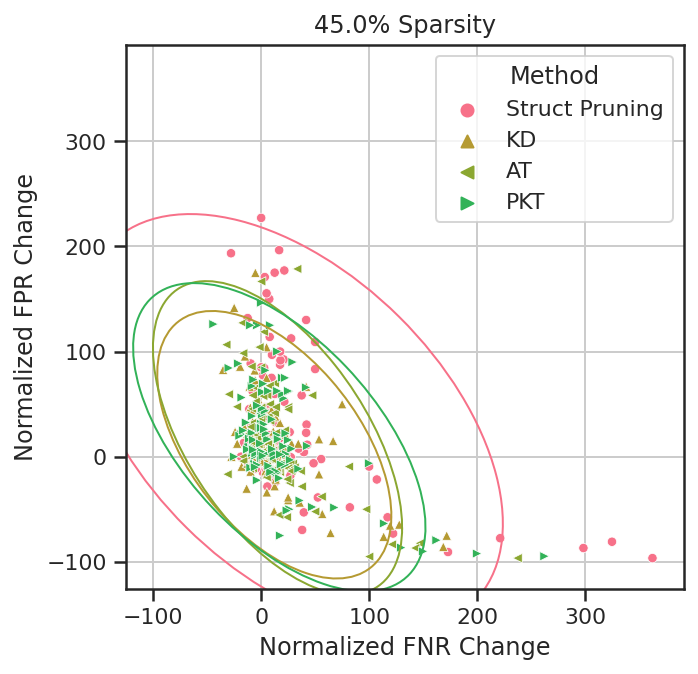}
        \caption{Pruning vs Distillation}
        \label{fig:PvFSP}
     \end{subfigure}
    \hfill
     \begin{subfigure}[b]{0.44\textwidth}
         \centering
         \includegraphics[width=\textwidth]{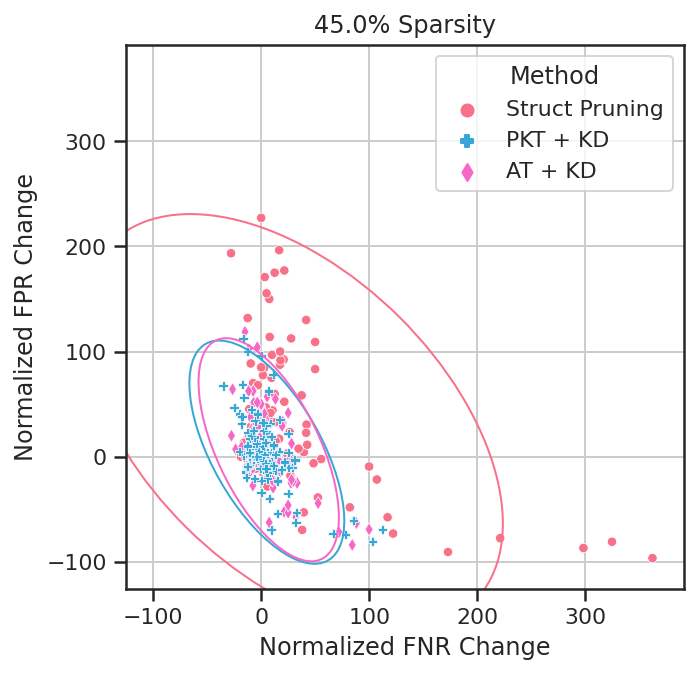}
        \caption{AT/PKT + KD}
        \label{fig:KDvCRD}
     \end{subfigure}

    \caption{Normalized FP/FN Rate Change for  distillation methods on biased CIFAR100 dataset}
    \label{fig:bias}
    
\end{figure}


While the results we have presented so far are encouraging, they are merely based on our experiments on CIFAR100 (a well balanced dataset), which is not necessarily reflective of datasets in the wild. Real world data is often unbalanced and could amplify the bias induced by pruning. Hooker et al. reported that the negative effects of compression are most observed on underrepresented groups \cite{hooker2020characterising}. Therefore, we design a series of experiments to verify if knowledge distillation can mitigate data induced bias in pruned networks. We resample the CIFAR100 dataset to purposely under represent certain classes in the training set. We select 15 classes and reduce the remaining training samples to 50\%, 20\%, and 10\% of their original numbers. Table \ref{tab:sampling} shows the selected classes and their original Top-1 accuracy. These classes are selected according to which classes from all our previous pruning experiments have either the smallest change in FPR/FNR or have an increase in accuracy. Our hypothesis is that these biased classes with less data samples will more likely be picked as "victims" by the pruned models.  We also test if combining our feature map based distillation methods with KD (e.g. AT + KD or PKT + KD) can achieve additional benefits.


From Table \ref{tab:sampling}, we can see that pruning alone performs significantly worse than knowledge distillation on nearly all classes with removed samples (marked in red). Noteably the "shrew" class went from 76.49\% top-1 accuracy when pruned on the entire dataset to just 29.99\% when reduced to just 10\% of of its original training examples. While AT + KD reached 54.24\% top-1 accuracy, an  88.09\% improvement. In addition, the best outcomes (bolded) for down sampled classes all came from combining distillation methods that use internal feature maps with KD. Table \ref{tab:similarity_tb} shows that Attention and PKT + KD reduced SDE in half at 45\% sparsity and CEV to less than 25\% of its value with pruning alone. Looking at the distribution of FPR/FNR change in Figure \ref{fig:bias}, we can see that the improvements are not limited to the 15 classes we intentionally down sampled. The models pruned normally have seen a substantial number of classes with significant shifts in both the false positive (FP) and false negatives (FN) direction. The pruned models over guess classes not down sampled and under guess those down sampled (i.e. bias is amplified). It is notable that the distributions of error change of PKT/AT + KD in Figure \ref{fig:bias} (b) are clearly more desirable than that of pruning alone pruning, even though they have nearly identical top-1 accuracy. This reinforces the usefulness of CEV/SDE in evaluating model bias. We conclude that knowledge distillation methods, especially those that align both feature maps as well as the output layer, are incredibly effective at mitigating pruning induced bias, even with unbalanced datasets.  




\section{Explaining Model Bias Using Model Similarity}

Previous studies \cite{blakeney2020pruning, movva2020dissecting} have shown that pruned neural networks evolve to substantially different representations while striving to preserve overall accuracy. In Section 3, we have demonstrated that knowledge distillation can effectively mitigate both pruning and data induced bias in compressed networks. Inspired by both work, we will be able to explain the occurrence of bias in pruned networks if strong correlations between model similarity and model bias can be confirmed. 

\begin{figure}[h]

\centering
\includegraphics[width=.45\textwidth]{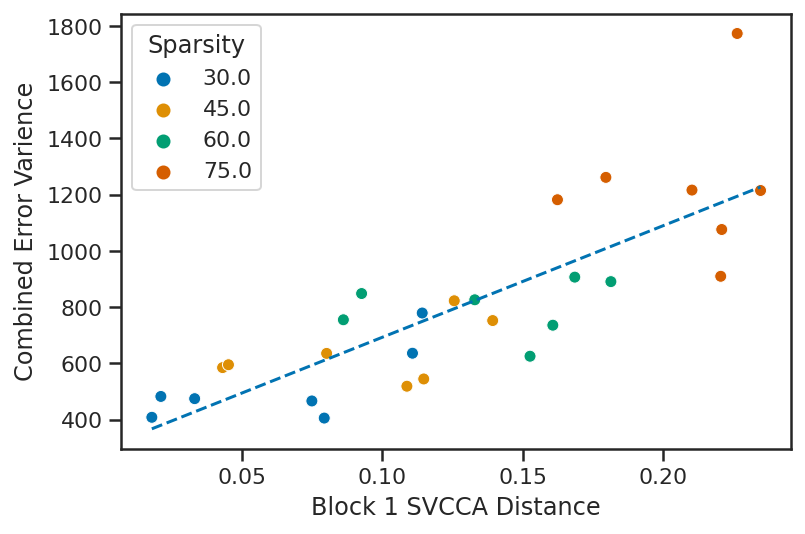}
\includegraphics[width=.45\textwidth]{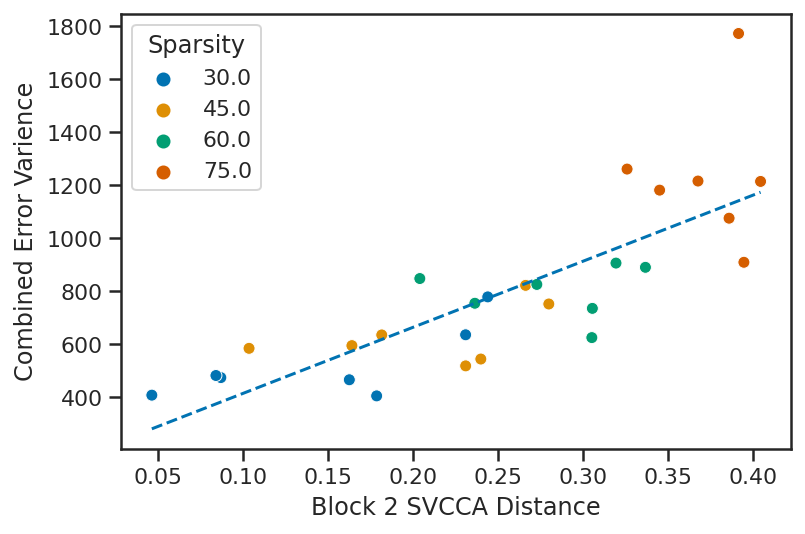}
\includegraphics[width=.45\textwidth]{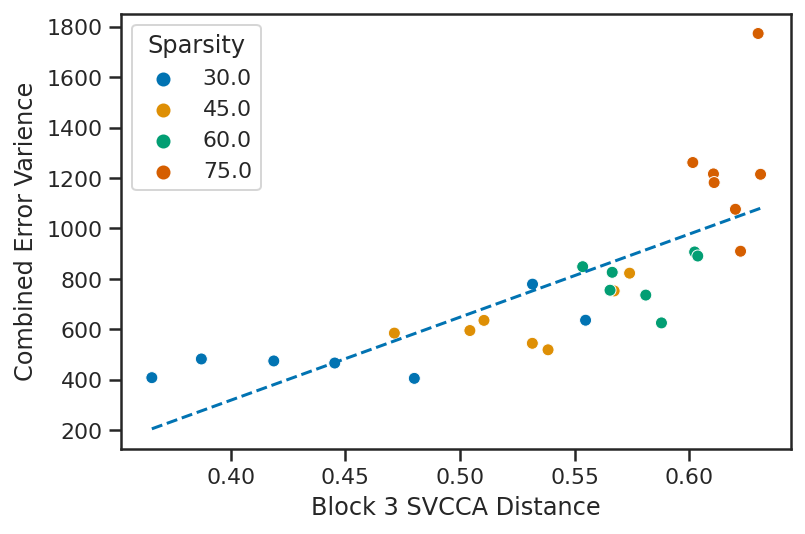}
\includegraphics[width=.45\textwidth]{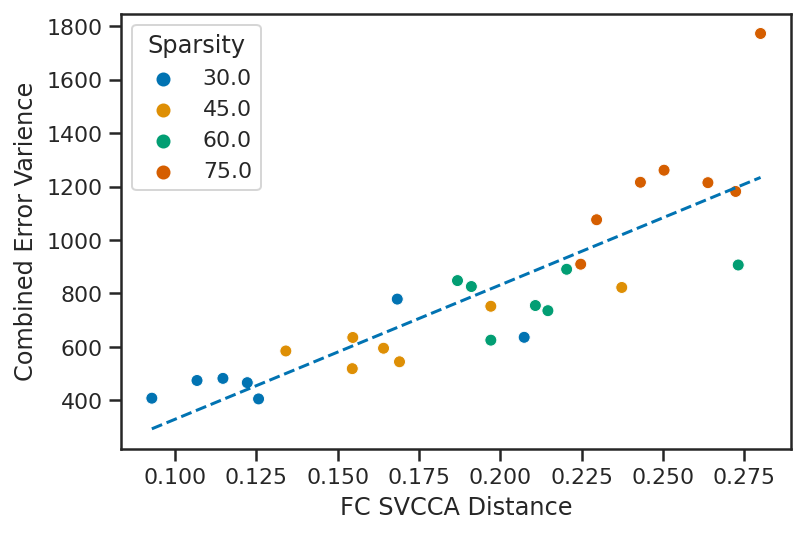}

\caption{Scatter Plot of Combined Error Variance Plotted against SVCCA Distance at each layer output with regression line overlaid. Values are colored by network sparsity}
\label{fig:sim_plot}
\end{figure}

\begin{table}[!ht]
\centering
\caption{Average SVCCA distances at each block and CEV/SDE results on biased CIFAR100 dataset}
\resizebox{.8\columnwidth}{!}{

\begin{tabular}{rlrrrrrrr}
\toprule
   \# PIEs &         Method &  Block 1 &  Block 2 &  Block 3 &    FC &  CEV &  SDE &  Accuracy \\
\midrule
742 &        AT + KD &    0.051 &    0.121 &    0.480 & 0.122 &                  851.694 &                    25.246 &    77.100 \\
748 &       PKT + KD &    0.071 &    0.164 &    0.495 & 0.128 &                  903.329 &                    25.103 &    77.335 \\
768 &        SP + KD &    0.107 &    0.231 &    0.538 & 0.161 &                 1506.103 &                    30.974 &    76.927 \\
742 &       FSP + KD &    0.063 &    0.187 &    0.514 & 0.156 &                 1518.462 &                    30.668 &    75.285 \\
770 &             KD &    0.106 &    0.230 &    0.540 & 0.160 &                 1536.566 &                    30.789 &    78.142 \\
909 &             AT &    0.044 &    0.104 &    0.475 & 0.142 &                 1955.926 &                    37.000 &    78.097 \\
881 &            PKT &    0.075 &    0.167 &    0.506 & 0.154 &                 2190.690 &                    38.121 &    78.963 \\
838 &             SP &    0.113 &    0.240 &    0.535 & 0.187 &                 2654.847 &                    41.378 &    78.520 \\
877 &            FSP &    0.045 &    0.162 &    0.508 & 0.171 &                 2901.940 &                    43.169 &    78.413 \\
887 & Struct Pruning &    0.127 &    0.268 &    0.581 & 0.261 &                 4237.391 &                    51.146 &    77.242 \\
\bottomrule
\end{tabular}}
\label{tab:similarity_tb}

\end{table}


We utilize the Singular Vector Canonical Correlation Analysis (SVCCA) \cite{raghu2017svcca} to study and contrast layer representations of pruned and original models. It combines Singular Vector Decomposition (SVD) to reduce the dimensionality of layer activations, and Canonical Correlation Analysis (CCA), to maximize a projection between two layers that results in the highest correlation. For a given dataset $X = \{x_1, ... , x_m\}$ and a neuron $i$ on layer $l$, the output of that neuron on the entire dataset is defined as the vector $z^{l}_i$. SVCCA takes input from two layers $l_1 = \{z^{l_1}_1, ... , z^{l_1}_{n_1}\}$ and $l_2 = \{z^{l_2}_1, ... , z^{l_2}_{n_2}\}$ where $n_1$ and $n_2$ are the number of neurons in their respective layers. SVD is performed on each layer to get new subspaces $l'_1$ and $l'_2$ where $l'_1 \subset l_1$, $l'_2 \subset l_2$. Next,  $l'_1$ and $l'_2$ are linearly transformed to be as aligned as possible and correlation coefficients are calculated. SVCCA outputs aligned directions $(\tilde{z}^{l_1}_i, \tilde{z}^{l_2}_i)$ and how well they correlate. The higher the correlation value $\rho_i$ is, the more similar the two aligned directions are. 

To calculate the similarity, we compare the feature maps and activations of the baseline model in our compressed populations. We use the same sets of features targeted by our various distillation methods. These activation output blocks in ResNet are the feature maps in the model where their input shapes change. In CIFAR100, the output shapes include 32x32, 16x16, 8x8. We also look at the similarity of the fully connected output layer before softmax, which is a tensor of size 100. In our experiments, we calculate the SVCCA distance  $1 - \bar{\rho}$ where $\bar{\rho}$ is the mean of the $\rho_i$ values from the top SVCCA similarity. A larger SVCCA distance indicates a lower similarity between two layers (or block of layers). 

The calculated SVCCA distance and corresponding CEV values at different sparsities are shown in Figure \ref{fig:sim_plot}, which are strongly correlated as evidenced by the regression line. In other words, a pruned model that yields higher similarity to the original model will have better CEV values (i.e. less induced bias). High similarity from the FC layer leading to better outcomes may intuitively be less surprising. After all, if two models with different configurations both reach the same answer, a small SVCCA distance is expected. However, we observe the strong correlations between similarity and CEV not only in the FC layer but \textbf{\textit{all}} levels of model features.

If knowledge distillation shows \textit{how} to reduce desperate impact of underrepresented classes when pruning, we believe network feature similarity explains \textit{why} it works. Regardless of sparsity level or pruning methods, we find the best predictor of CEV in our models is SVCCA layer distance. This holds true for both our experiments with blanced CIFAR100 and biased CIFAR100, as shown in Table \ref{tab:similarity_tb}. Additionally, we find that top-1 accuracy is not a great indicator of CEV. In fact, our models with the highest top-1 accuracy on the biased dataset are ranked near the bottom of CEV/SDE scores. We believe the strong correlation between similarity and CEV provides an important direction for future research in developing better pruning and distillation methods with mitigated bias. 

     
     

    

\section{Related Work}

With the ubiquitous usage of AI, a substantial literature has recently emerged concerning algorithmic bias, discrimination, and fairness. Here we only discuss relevant studies that focused on addressing bias induced by data and algorithms. Mehrabi et al. conducted a survey on bias and fairness in machine learning \cite{mehrabi2019survey}. Mitchell et al. explored how model cards can be used to provide details on model performance across cultural, racial, and intersectional groups and to inform when their usage is not well suited \cite{mitchell2019model}. Gebru el al. proposed to use datasheets as a standard process for documenting datasets \cite{gebru2018datasheets}. Amini,et al. proposed to mitigate algorithmic bias through resampling datasets by learning latent features of images \cite{amini2019uncovering}. Wang et al. designed a visual recognition benchmark for studying bias mitigation in visual recognition. \cite{wang2020towards}. Literature on network pruning has been historically focused on accuracy \cite{lecun1990optimal, evci2020rigging} with recently work on robustness \cite{guo2018sparse, wang2018adversarial}. Movva and Zhao \cite{movva2020dissecting} investigated the impact of pruning on layer similarities of NLP models using LinearCKA \cite{kornblith2019similarity}. Ansuini et al. and Blakeney et al. also investigated how pruning can change representations using similarity based measures \cite{ansuini2020similarity, blakeney2020pruning}. Unfortunately, very few work have studied how pruning can induce bias and how to evaluate and mitigate the pruning induced bias. To the best of knowledge, our work is the first one to tackle the evaluation, mitigation and explanation of pruning induced bias altogether. 

\section{Broader Impact}


Compressed DNNs are widely used and our research can help alleviate increasing concerns in society regarding algorithmic bias in AI. Specifically, (1) we help the community better understand why bias occur in pruned models. (2) we present a viable solution to mitigate algorithmic bias induced by pruning. (3) CEV and SDE provide new metrics to evaluate the bias-prevention quality of pruned models. However, our research is currently limited to study and mitigate bias induced by pruning only. We believe there are many other types of biases exist in various AI models, which may not be addressed by this work. Our metrics are not a replacement for carefully studying the behavior of models and ensuring that they meet other fairness standards. Therefore, we do not encourage policy makers or stakeholders use our metrics to establish a scoring system for ranking general AI models. 


\section{Limitations}

Our work has three primary limitations. First, we only show extensive experimental results on a single model (ResNet) and dataset(CIFAR100). However, we do include limited eexperiments using VGG16 on CIFAR10, Imagenette, and Imagewoof in the appendix. Second, our model population size for each method/sparsity is much lower than the populations size (30 each) of Google's PIE paper \cite{hooker2019compressed}. While more results could increase our confidence, we believe our experiments are sufficient to draw conclusions from. Third, there is no human involvement in our method. The combination of automatic bias prevention techniques (like ours) and human-in-the-loop process (e.g. auditing) may further reduce biases in pruned neural networks. 





\section{Conclusion}

In this paper, we propose two new metrics, Combined Error Variance (CEV) and Symmetric Distance Error (SDE), to evaluate the pruning induced bias in compressed neural networks. We also demonstrate that knowledge distillation can effectively mitigate bias in pruned models, even with unbalanced datasets. Moreover, we reveal that model similarity has strong correlations with pruning induced bias, which can be used to explain why bias occurs in pruned neural networks.


\bibliographystyle{IEEEtranSN}
\bibliography{ref}

\appendix

\section{Appendix}

In this appendix, we include an additional set of experiments that are all conducted on VGG16 with FSP distillation using various datasets. Specifically, Figures 1, 2, and 3 show the Accuracy, PIEs, CEV, and SDE results for the CIFAR10, Imagenette, Imagewoof dataset respectively. Although these results are not as complete as our ResNet experiments on CIFAR100 (because only one distillation method is evaluated), they do provide strong complementary evidences that enhance two primary findings and conclusions made in the main paper: (1) CEV and SDE can better reflect pruning induced bias; and (2) Knowledge distillation can mitigate induced bias in pruned neural networks without compromising accuracy. 




\begin{figure*}[!h]
    \centering
    \includegraphics[width=.80\textwidth]{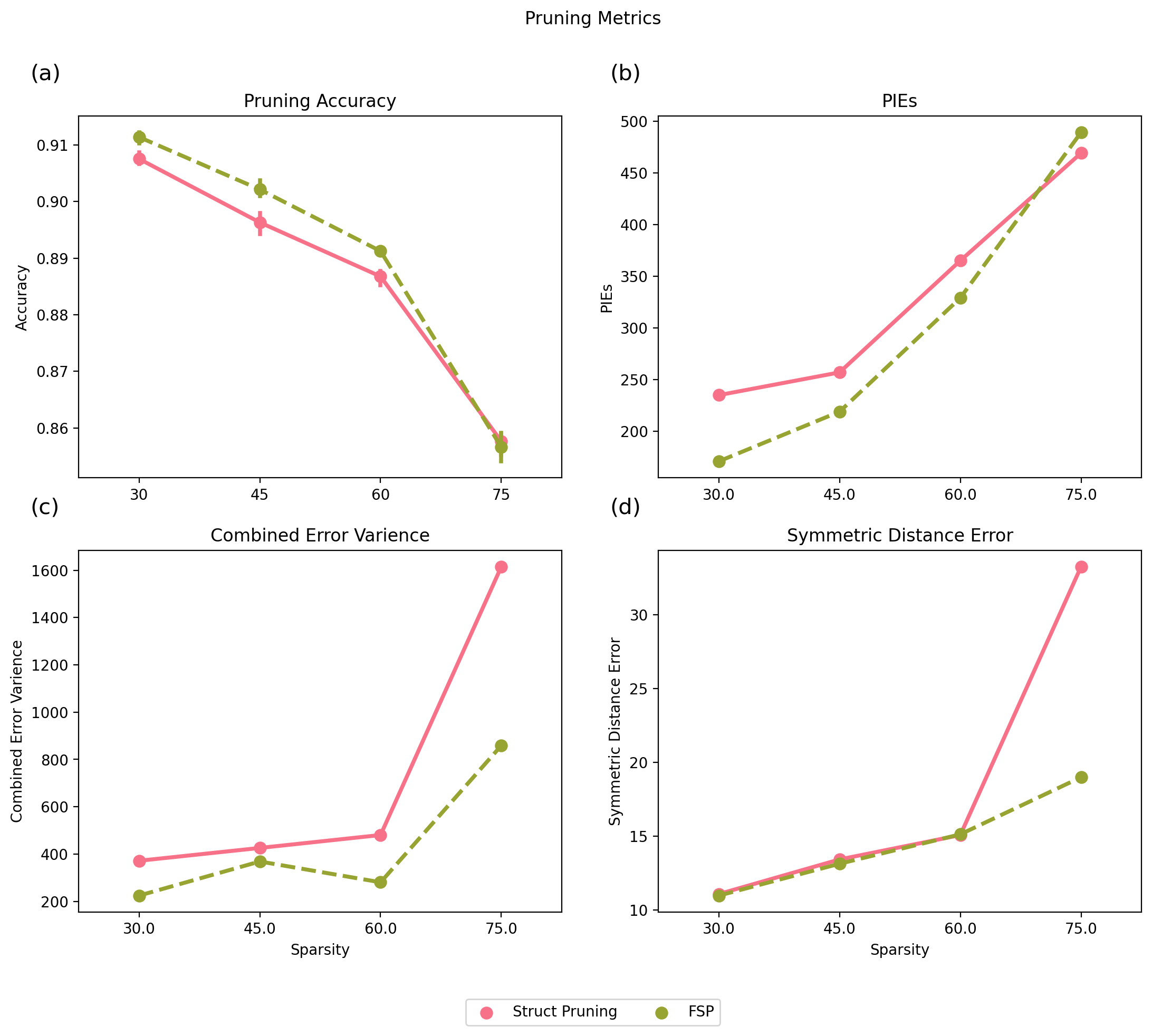}

    \caption{Accuracy, PIEs, CEV, and SDE results for CIFAR10}
    \label{fig:join_cifar}
\end{figure*}

\begin{figure*}[!h]
    \centering
    \includegraphics[width=.80\textwidth]{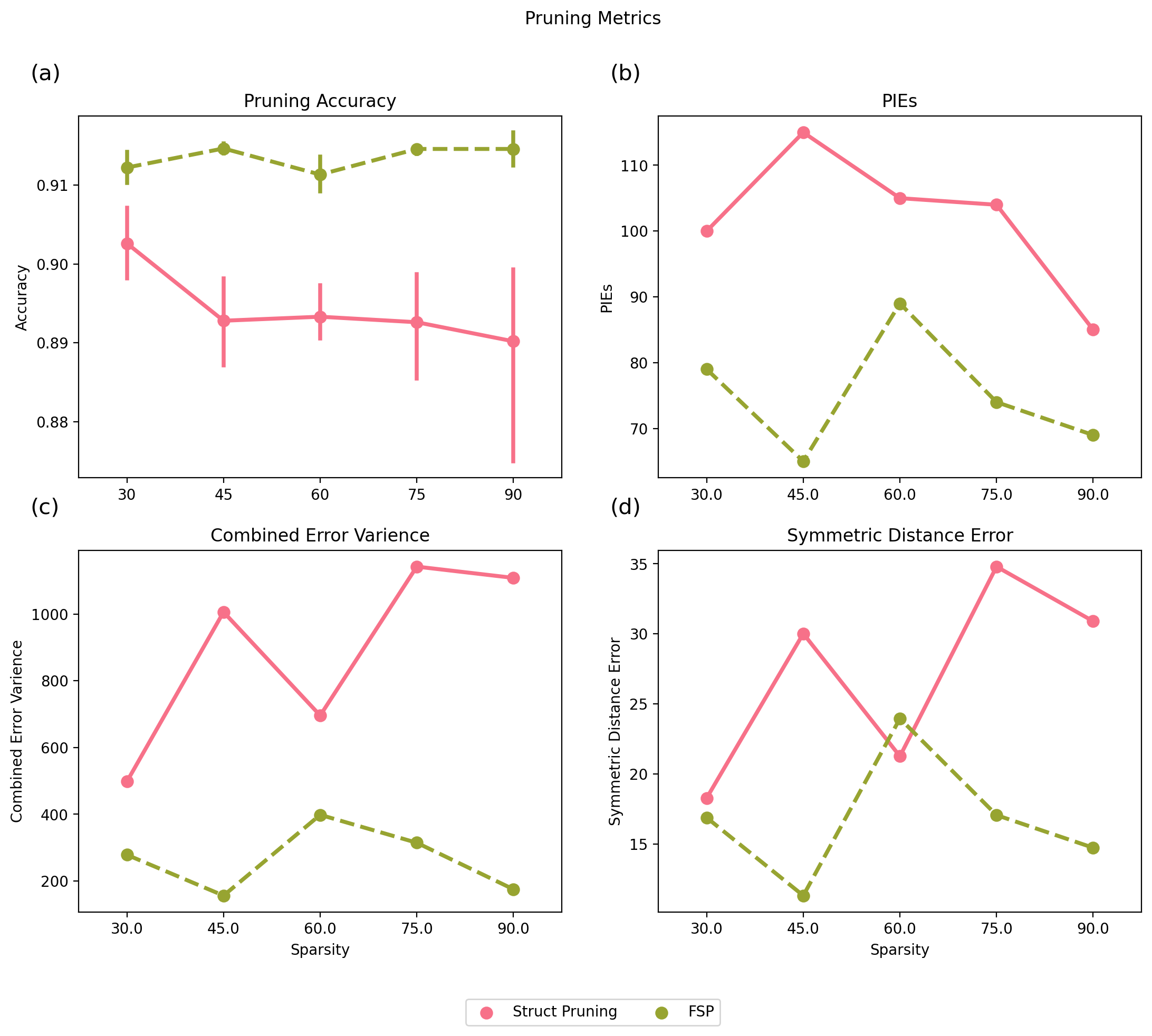}

    \caption{Accuracy, PIEs, CEV, and SDE results for Imagenette}
    \label{fig:joint_imagenette}
\end{figure*}

\begin{figure*}[!h]
    \centering
    \includegraphics[width=.80\textwidth]{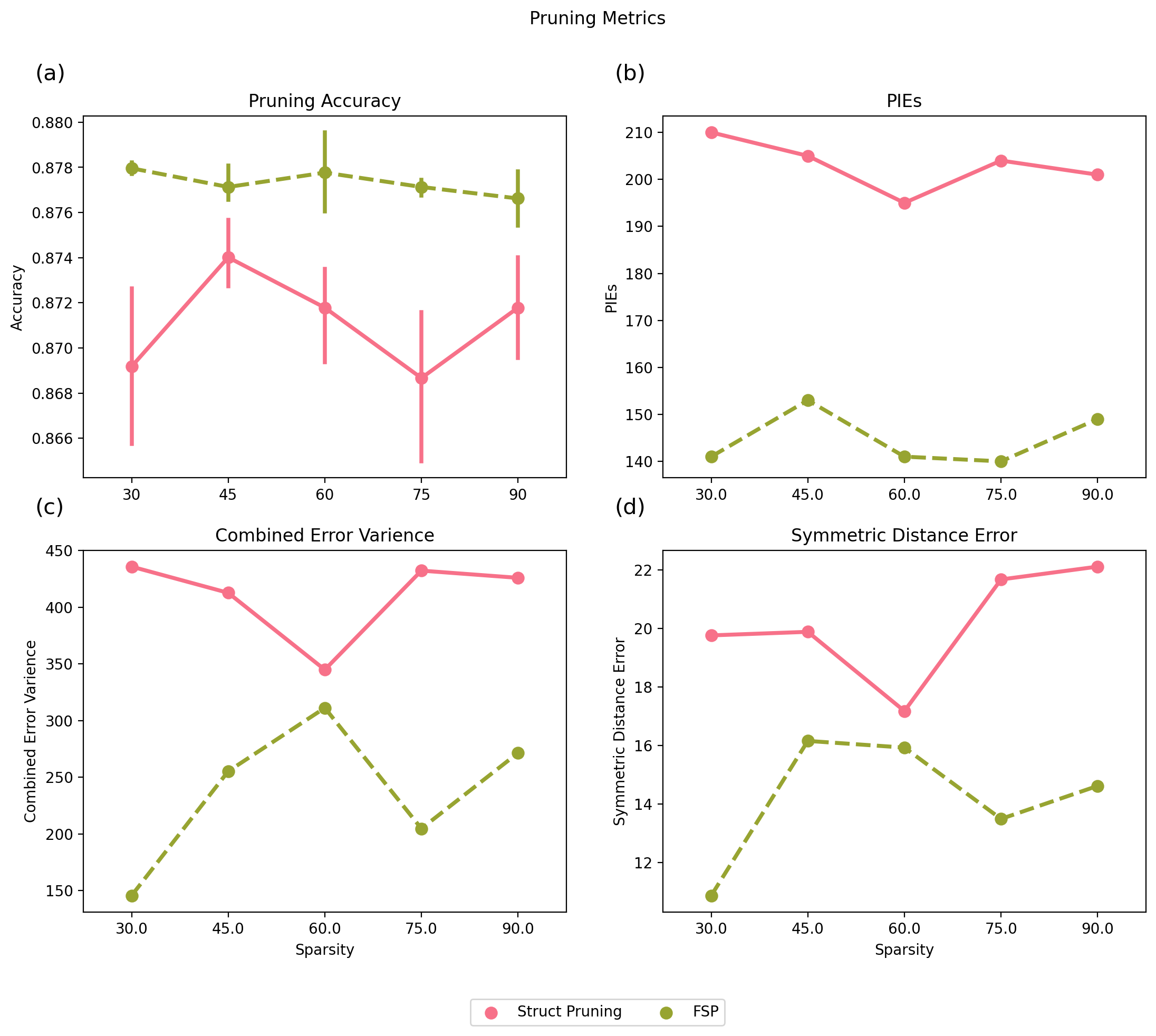}

    \caption{Accuracy, PIEs, CEV, and SDE results for Imagewoof}
    \label{fig:joint_imagewoof}
\end{figure*}

\begin{figure}[h]

\centering
\includegraphics[width=.49\textwidth]{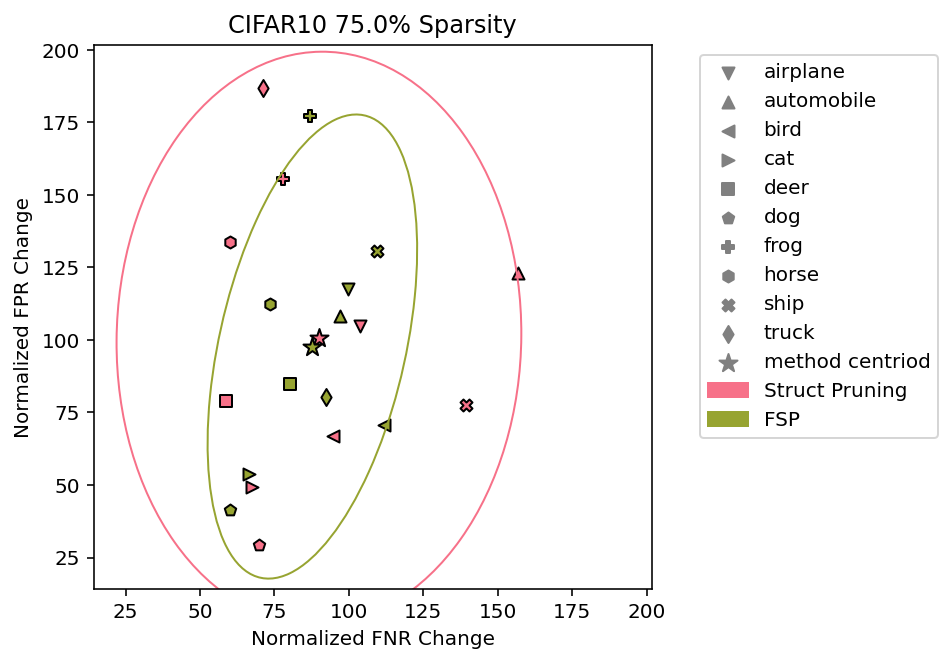}
\includegraphics[width=.49\textwidth]{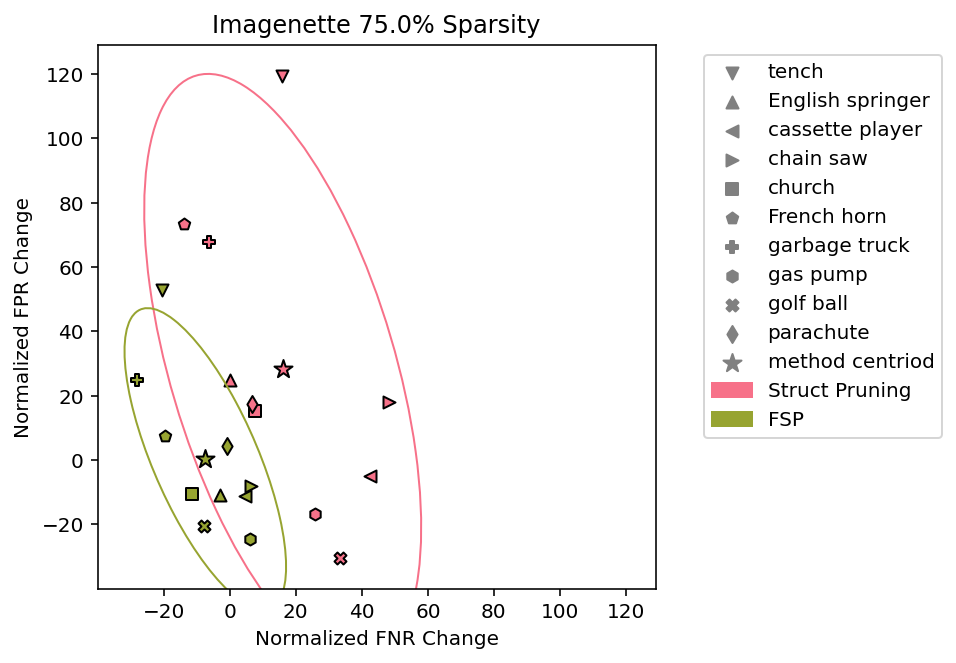}
\includegraphics[width=.49\textwidth]{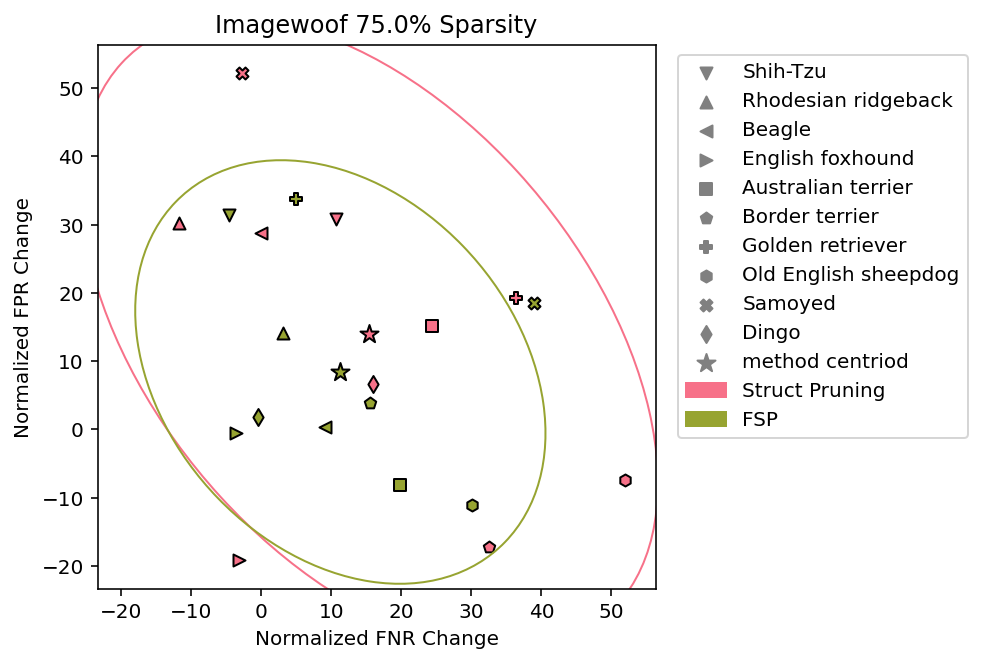}
\includegraphics[width=.49\textwidth]{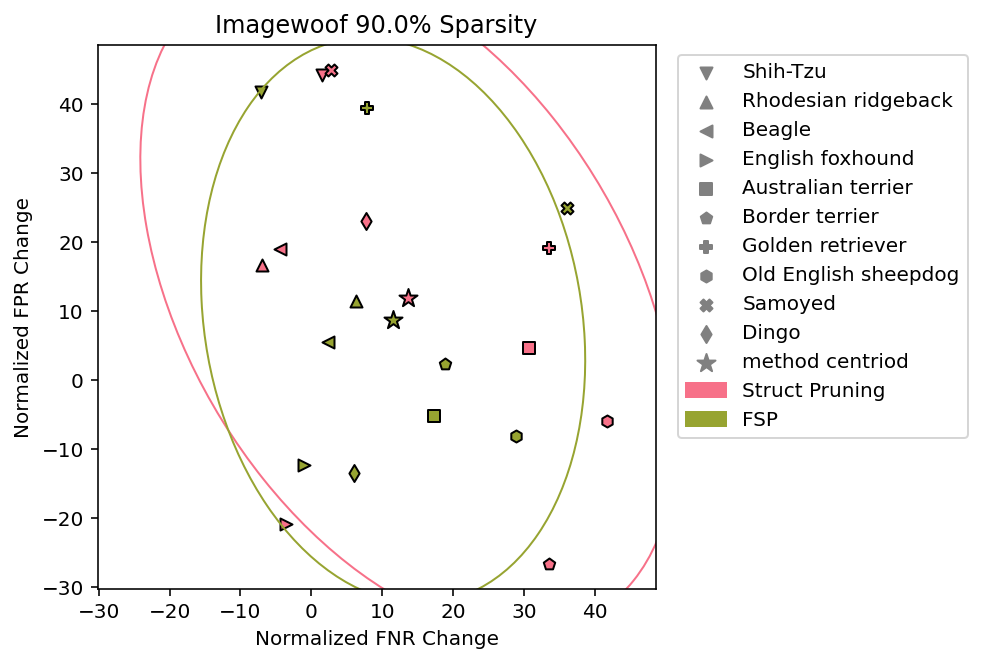}

\caption{Normalized FNR/FPR change of VGG16. Here ellipses represents 2 standard deviations from the mean. We have also plotted the mean class change as the star-shaped 
``method centriod''.}
\label{fig:sim_plot_vgg}
\end{figure}

\end{document}